%% file: main.tex
\documentclass[letterpaper,10pt,conference]{ieeeconf}

\IEEEoverridecommandlockouts
\usepackage{amsmath}
\usepackage{amssymb}
\usepackage{booktabs}
\usepackage{tabularx}
\usepackage{array}
\usepackage{multirow}
\usepackage{placeins}
\usepackage{capt-of}
\usepackage{graphicx}
\usepackage{microtype}
\usepackage[dvipsnames]{xcolor}
\usepackage{tikz}
\usetikzlibrary{arrows.meta}
\usepackage[font=small]{caption}
\usepackage[hidelinks]{hyperref}
\hypersetup{%
    pdftitle={Before We Act: Learning Action-Conditioned Team-Response Beliefs for Decentralized Multi-Robot Manipulation},
    pdfsubject={arXiv preprint},
    pdfkeywords={multi-robot manipulation, decentralized coordination, team-response beliefs, CARE}%
}

\newcommand{\nrob}[1]{$_{\text{#1}}$}
\newcommand{\nr}{\textemdash}

\newcommand{\studytable}{%
    \footnotesize\setlength{\tabcolsep}{4pt}%
    \renewcommand{\arraystretch}{1.0}%
}
\newcommand{\studynote}[1]{%
    \par\vspace{0.5pt}{\scriptsize\raggedright #1\par}%
}

\AtBeginDocument{%
  \setlength{\abovedisplayskip}{3pt plus 1pt minus 1pt}%
  \setlength{\belowdisplayskip}{3pt plus 1pt minus 1pt}%
  \setlength{\abovedisplayshortskip}{1pt plus 1pt}%
  \setlength{\belowdisplayshortskip}{2pt plus 1pt minus 1pt}%
}

\newcommand{\method}{\textsc{Care}}
\newcommand{\E}{\mathbb{E}}

\newcommand{\Law}{\mathcal{L}}

\title{\LARGE \bf
Before We Act: Learning Action-Conditioned Team-Response Beliefs \\ for Decentralized Multi-Robot Manipulation
}

\input{authors}

\IEEEaftertitletext{%
\begin{minipage}{\textwidth}
    \centering
    \includegraphics[width=\textwidth]{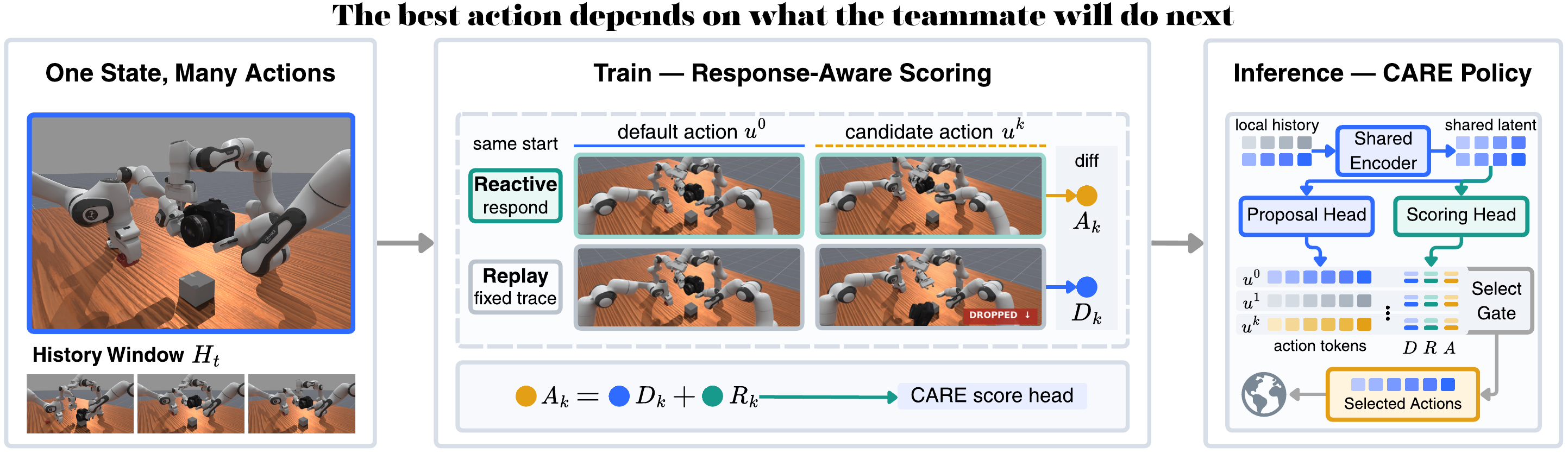}
    \vspace{-1.5em}
    \captionof{figure}{\textbf{Before acting, predict what teammate response
    contributes.} From the same local history (left), CARE compares candidate
    and nominal actions under reactive and replayed teammate behavior during
    training (center). The difference between total advantage $A_k$ and direct
    advantage $D_k$ supplies the response label $R_k$. At execution (right),
    the response readout uses predictive memory to guide selection, with
    calibrated fallback to the nominal action.}
    \label{fig:care-teaser}
\end{minipage}%
\vspace{0.2\baselineskip}%
\vspace{-0.5em}
}

\begin{document}

\maketitle
\thispagestyle{plain}
\pagestyle{plain}

\begin{abstract}
In multi-robot manipulation, an action's value depends on how teammates respond. We present \method{} (Cooperative Action--Response Belief Estimation), a framework for learning \emph{team-response beliefs} from local history. A predictive belief encoder represents the interaction and conditions a nominal action. Matched reactive and replay continuations separate each candidate's direct advantage from the utility contributed by teammate adaptation. A candidate-conditioned readout learns this response distribution alongside total advantage, and calibrated selection chooses an alternative or retains the nominal action. Our analysis characterizes when response changes the preferred action and how prediction and source errors limit the recoverable gain. With the representation, proposals, and data fixed, explicit response supervision improves candidate ranking and reduces selection regret relative to learning the same information through branch utilities. The benefit is concentrated in decisions where teammate adaptation changes the preferred action. Across RoboFactory, the MARS Control Track, and physical collaboration tasks, CARE achieves the highest average success among the evaluated policies, combining a capable nominal policy with selective coordination adjustments. Together, these results show that CARE turns predictions of teammate adaptation into effective decentralized action selection. Project page: \url{https://jeong-zju.github.io/care/}.
\end{abstract}

\section{Introduction}

Consider a robot carrying an object with a partner and deciding when to turn. Turning now shortens its own path but can destabilize the object if the partner is still regrasping, whereas waiting briefly makes little direct progress but lets the partner finish the grasp. The uncertainty that matters is how the team will respond to the robot's next action. The same decision arises when releasing an object in a handoff, yielding a shared workspace, or synchronizing a lift.

We define a \emph{team-response belief} as the distribution of a candidate's gain or loss in team utility due to teammate adaptation, relative to replaying nominal teammate commands. As in partially observable control~\cite{kaelbling1998pomdp}, this belief summarizes uncertainty from local history. Latent teammate modeling \cite{he2025latenttom}, scene representations \cite{wang2025gaudp}, shared policies \cite{doshi2026chorus}, language negotiation \cite{mandi2023roco}, and memory \cite{li2026trace} provide complementary tools for coordination. Our focus is the action-conditioned utility contrast and its prediction from local history.

Demonstrations train capable imitation policies \cite{zhi2022diffeomorphic,zhao2023act,chi2023diffusion,yan2025maniflow,liu2024rdt,kim2025oft,black2025pi05}, but record only the action taken and the response that followed. They do not specify how teammates would have responded to alternatives. Multi-agent critics \cite{lowe2017maddpg,iqbal2019aac,zhang2025defmarl} can learn their combined value without separating teammate adaptation. CARE compares reactive and replayed continuations to measure whether this adaptation helps or hurts team utility. Our hypothesis is that explicitly supervising this contribution improves finite-data learning when response changes the preferred action.

\method{}, short for \emph{Cooperative Action--Response Belief Estimation}, connects predictive representation learning with \emph{paired interaction augmentation}. Its \emph{predictive belief encoder} represents the evolving interaction from local history, supporting both nominal action generation and response inference. Paired continuations compare each candidate with the nominal action under reactive teammates and teammates replaying nominal commands. Subtracting the replay advantage from the reactive advantage isolates the response interaction. A candidate-conditioned response model connects this supervision to the predictive representation. We retain the original demonstrations and add explicit interaction supervision, complementing methods that reshape demonstrations \cite{chen2026sai,zhao2026triman,zhao2026nestdex}.

We train CARE in two stages. We first learn the predictive belief encoder and its action interface on demonstrations, then freeze them as the reference policy $\pi_{\mathrm{ref}}$. A fixed paired dataset trains the response readout, whose total-advantage lower quantile is calibrated for selection \cite{romano2019cqr,angelopoulos2021conformal,dabney2018iqn}. This separation lets us compare response targets with total-value and information-matched branch-utility targets while holding representation, proposals, and data fixed. Closed-loop success separately evaluates the complete policy $\pi_{\mathrm{CARE}}$. Its selector-off control retains the predictive encoder and measures the nominal path.

This paper makes three contributions.
\begin{itemize}\setlength{\itemsep}{0pt}
    \item We formulate action-conditioned team-response beliefs through matched reactive and replay contrasts and learn them from local predictive memory.
    \item We characterize when response changes the preferred action and how source and prediction errors limit the gain available to a learned selector.
    \item We demonstrate the finite-data benefit of response supervision with fixed-proposal, information-matched comparisons, alongside complete-policy evaluation in simulation and on hardware.
\end{itemize}

\begin{figure*}[t]
    \centering
    \input{figure_care_labels}
    \caption{\textbf{CARE learns and acts on team-response beliefs.} A predictive belief encoder represents the interaction from local history and conditions the nominal action (left). Paired reactive and replay contrasts distinguish teammate adaptation from direct effects through $R=A-D$ (center). A candidate-conditioned readout estimates these advantages from the learned representation, which remains fixed during response learning. Calibrated selection chooses a supported alternative or retains $u^0$ (right). Execution requires only local history.}
    \label{fig:care-method}
    \vspace{-0.5em}
\end{figure*}
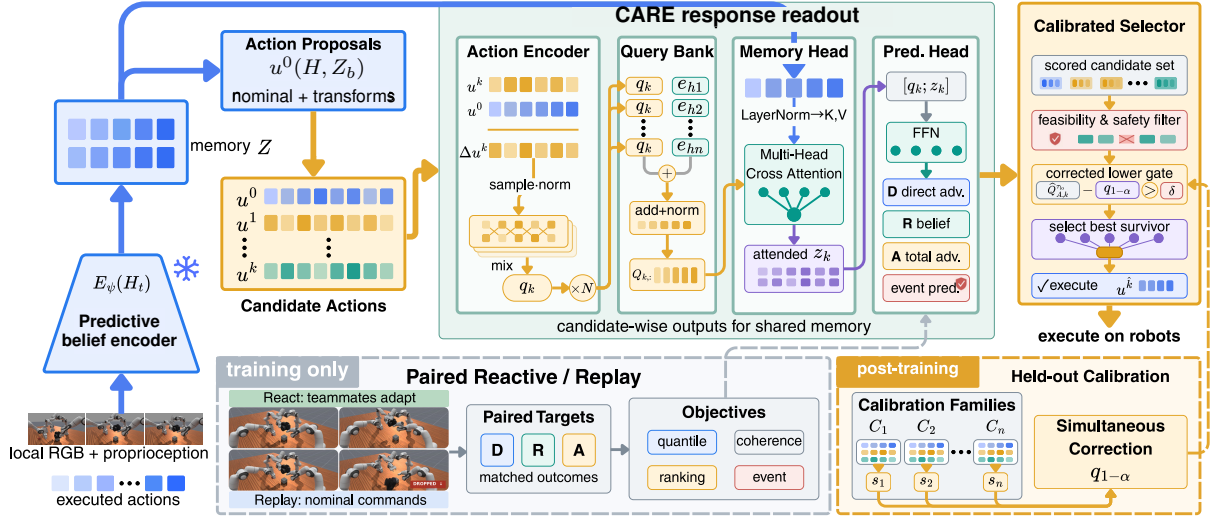

\section{Learning Predictive Team Beliefs}
\label{sec:belief}

\subsection{Defining Team-Response Belief}

Robot $i$ acts from local history $H=H_t^i=(o_{\leq t}^i,u_{<t}^i)$, including its RGB observations, proprioception, executed actions, and task instruction. CARE constructs a finite candidate set $\mathcal U(H)=\{u^0,\ldots,u^{K-1}\}$ around its nominal action chunk $u^0$. Alternatives such as waiting, retreating, or slowing down use only the acting robot's chunk and joint state (Section~\ref{sec:experiments}). Let $U(Y)$ be a fixed bounded team utility over a given horizon, specified experimentally in Section~\ref{sec:experiments}. Candidates, utility, horizon, and reference teammates define the response target.

\subsection{Predictive Belief Encoder}
\label{sec:predictive-encoder}

An instantaneous local observation may be compatible with different stages of a teammate's motion. CARE therefore learns a predictive latent belief from history, so that action generation and response estimation can depend on temporal evidence. The \emph{predictive belief encoder} parameterizes
\begin{equation}
p_\phi(z\mid H)=\prod_{j,f}\operatorname{Cat}\bigl(z_{jf};p_{\phi,jf}(H)\bigr),
\label{eq:predictive-belief}
\end{equation}
where $j$ indexes interaction slots and $f$ categorical factors. This learned distribution represents predictive alternatives without assigning semantic labels to hidden teammate intent. Let $Z_b(H)$ denote a continuous embedding of its categorical probabilities and $Z_e(H)$ retained evidence of past prediction discrepancies. Their combined representation $Z(H)=[Z_b(H);Z_e(H)]$ supports response inference. To make the belief relevant to control, CARE conditions the nominal action on $Z_b$ as
\begin{equation}
u^0(H)=G_\theta(H)+g(H)\odot\Delta_\theta(H,Z_b).
\label{eq:belief-proposal}
\end{equation}
Here $G_\theta$ is a temporal imitation policy, $\Delta_\theta$ a belief-conditioned correction, and $g\in[0,1]$ an evidence-dependent gate. The correction allows the same local observation to yield different actions when its preceding interaction differs.

Learning makes the belief informative about demonstrated actions and subsequent interaction, using team and future evidence available only during training (Section~\ref{sec:training-objectives}). Predicting demonstrated interactions alone does not identify responses to untried actions, motivating paired supervision.

\subsection{Paired Interaction Augmentation}
\label{sec:augmentation}

To distinguish an action's direct effect from the response it elicits, CARE compares two teammate behaviors under a fixed reference policy $\pi_{\mathrm{ref}}(H)=u^0(H)$. Let $\bar u^{-i}$ be the teammate command sequence induced by the nominal ego action. For each candidate $u^k$, define $Y_k^{\mathrm R}$ as the outcome when teammates respond according to $\pi_{\mathrm{ref}}$, and $Y_k^{\mathrm P}$ as the outcome when they replay $\bar u^{-i}$. The continuations share the initial interaction context and coupled exogenous randomness. Replay holds teammate commands fixed while preserving physical evolution. The paired advantages are
\begin{equation}
\begin{aligned}
D_k &= U(Y_k^{\mathrm P})-U(Y_0^{\mathrm P}),\quad
A_k = U(Y_k^{\mathrm R})-U(Y_0^{\mathrm R}),\\
R_k &= A_k-D_k .
\end{aligned}
\label{eq:decomposition}
\end{equation}
Here $D_k$ measures direct advantage with teammate commands fixed, and $R_k$ measures the additional gain or loss associated with their adaptation. Equivalently, $R_k$ compares the reactive--replay utility gap of the candidate with that of the nominal action. Physical interaction remains in both branches, with direct advantage defined by holding teammate commands fixed. The components may be dependent and are defined relative to the reference teammates and nominal trace. Simulator or model-generated labels must preserve this matched intervention. Altering observations alone does not identify $R_k$.

The \emph{team-response belief} for candidate $k$ is $b_H^k=\Law(R_k\mid H)$, with $u^k\in\mathcal U(H)$ fixed by the proposal rule. It describes uncertainty about response utility, whereas $p_\phi(z\mid H)$ represents predictive interaction state. A candidate-conditioned model estimates the former from the latter's representation. CARE predicts direct, response, and total quantiles separately because dependent marginal quantiles do not generally add.

\subsection{Why Representation and Pairing Are Complementary}
\label{sec:identification}

Two teammate policies may agree on demonstrated actions yet respond oppositely to a new candidate. Paired branches define a response target beyond those observations, while predictive memory supplies temporal evidence for estimating it locally. The readout estimates quantiles conditional on $(Z(H),u^k,u^0)$, so the representation must retain the temporal evidence needed to predict response utility. Predicting all four branch utilities provides the same label information as their contrasts. Directly learning $R_k$ directs shared features toward the utility of adaptation, making this contribution explicit during training. Q4 tests this learning bias with both the representation and proposals fixed.

\subsection{Decision Value of Response Information}
\label{sec:decision-value}

To characterize when teammate response changes the preferred action, let $d_k=\E[D_k\mid H]$, $r_k=\E[R_k\mid H]$, and $a_k=d_k+r_k$, with $d_0=r_0=a_0=0$. Under a fixed tie-breaking rule let $k_D\in\arg\max_k d_k$ and $k_*\in\arg\max_k a_k$. We define the decision value of response information as
\begin{equation}
\begin{aligned}
\Gamma(H)\!=\!a_{k_*}\!-\!a_{k_D}
=\max_k\bigl\{(r_k-r_{k_D})\!-\!(d_{k_D}-d_k)\bigr\}\!\geq\!0.
\end{aligned}
\label{eq:response-threshold}
\end{equation}
Response changes the optimal value exactly when some candidate's relative response gain exceeds its direct disadvantage. Waiting is useful, for example, when the partner's additional progress outweighs its direct cost. A large $|r_k|$ alone is not enough, nor is a reversal between two inferior candidates. Equation \eqref{eq:response-threshold} defines the strata of Q2.

\textbf{Recoverable decision value.} At a fixed history, let $\beta(H)$ bound the candidate-wise Wasserstein distance between source and target laws of $(D_k,R_k)$ under the $\ell_1$ transport cost. Suppose estimated component means differ from their source means by at most $\epsilon_D$ and $\epsilon_R$, uniformly over candidates. Since $(D,R)\mapsto D+R$ is 1-Lipschitz, $\widehat a_k=\widehat d_k+\widehat r_k$ obeys $|\widehat a_k-a_k|\leq e$, where $e=\beta+\epsilon_D+\epsilon_R$. The mean-based choice $\widehat k\in\arg\max_k\widehat a_k$ then satisfies
\begin{equation}
a_{\widehat k}(H)-a_{k_D}(H)\geq\Gamma(H)-2e(H),
\label{eq:belief-gain}
\end{equation}
Indeed, $a_{k_*}\leq\widehat a_{k_*}+e\leq\widehat a_{\widehat k}+e\leq a_{\widehat k}+2e$. Subtracting $a_{k_D}$ gives \eqref{eq:belief-gain}. A unique best candidate is recovered when its lead exceeds $2e$, and $\Gamma>2e$ ensures improvement over direct-only choice. Thus response magnitude matters through its effect on the decision margin. The same bound applies to an equally accurate total-value predictor. The benefit of decomposition lies in learning accurate decision scores from finite paired data, as examined in Q4.

\textbf{Reference teammates and deployment.} Freezing $\pi_{\mathrm{ref}}$ fixes its parameters, not its actions. Reactive teammates still act on the observations induced by each candidate, so paired labels capture adaptation through the nominal policy. At deployment, teammates can also select alternatives. These interventions and the histories they induce create the additional shift. When a selector abstains, it executes exactly $\pi_{\mathrm{ref}}$ at its current history. The shared nominal policy anchors deployment, although earlier interventions can change the history. Concurrent overrides can also reinforce or counteract one another beyond the response represented in the labels. Thus CARE evaluates candidates against reference teammates without solving for mutually consistent selector decisions. These differences enter $\beta$ alongside outcome-model mismatch. Equation~\eqref{eq:belief-gain} describes their effect on mean-based decisions.

\section{From Belief to Action}
\label{sec:acting}

\subsection{Action-Conditioned Response Readout}

The response model must distinguish different actions in the same interaction. CARE therefore conditions its advantage predictions jointly on the predictive representation and candidate as
\begin{equation}
\widehat Q_{X,k}^\tau(H)=f_{\omega,X}^\tau\bigl(Z(H),u^k,u^0\bigr),
\quad X\in\{D,R,A\}.
\label{eq:belief-readout}
\end{equation}
Here $\tau\in\mathcal T$ indexes quantile levels. Response quantiles summarize $\widehat b_H^k$, while total-advantage quantiles support conservative choice. Shared readout features let response supervision inform total-value estimation without imposing quantile additivity. An additional probability estimate describes adverse events. Training holds $Z$ and the proposal rule fixed (Section~\ref{sec:training-objectives}).

\subsection{Calibrated Selection under Uncertainty}
\label{sec:calibration}
\label{sec:selection}

After freezing the response readout, we calibrate its lower quantile on $M_{\mathrm{cal}}$ families, where family $j$ contains all candidates and matched repeats $\ell\in\mathcal R_j$ of one snapshot with paired advantages $A_{j,k,\ell}$ and $A_{j,0,\ell}=0$. For a lower level $\tau_{\mathrm{lo}}\in\mathcal T$, the family score
\begin{equation}
s_j=\max_{k>0,\;\ell\in\mathcal R_j}
\bigl[\widehat Q_{A,k}^{\tau_{\mathrm{lo}}}(H_j)-A_{j,k,\ell}\bigr]_+
\label{eq:calibration-score}
\end{equation}
is the largest overshoot in a family. Set $q_{1-\alpha}$ to the $\lceil(M_{\mathrm{cal}}+1)(1-\alpha)\rceil$-th smallest score, or $+\infty$ if this index exceeds $M_{\mathrm{cal}}$. The calibrated bound $\underline A_k(H)=\widehat Q_{A,k}^{\tau_{\mathrm{lo}}}(H)-q_{1-\alpha}$ shifts all candidates equally, preserving their ordering while raising the threshold for intervention. Among alternatives with predicted adverse-event probability at most $\eta$, the policy picks $\widehat k\in\arg\max_k\underline A_k(H)$ and executes
\begin{equation}
\pi_{\mathrm{CARE}}(H)=
\begin{cases}
u^{\widehat k}, & \underline A_{\widehat k}(H)>\delta,\\
u^0, & \text{otherwise},
\end{cases}
\label{eq:selector}
\end{equation}
keeping $u^0$ when no alternative is feasible. The margin $\delta$ raises the evidence required to leave the nominal action and the threshold $\eta$ shrinks the set of actions considered, and both are fixed before any closed-loop evaluation.

\textbf{Role of calibration.} For a frozen readout and exchangeable calibration and test families, the order-statistic construction covers all candidates and repeats of a new family with probability at least $1-\alpha$. The maximum accounts for selection among one robot's candidates, allowing dependence within a family. It does not cover joint interventions by multiple selectors. Coverage is marginal over families from the reference distribution, and teammate interventions or hardware shift can break exchangeability. Q5 evaluates calibration through task success and intervention frequency during collective execution.

\subsection{Training Objectives}
\label{sec:training-objectives}

\textbf{Predictive representation.} Imitation supervises nominal actions, while prediction of future features and teammate behavior shapes the belief. Categorical KL alignment transfers predictive structure from a teacher with team observations and future evidence to local inference. Separate gradients update each distribution, while teacher reconstruction preserves future information. Prediction targets and regularization are specified below.

\textbf{Response learning.} With the encoder and nominal policy fixed, paired advantages train only the response readout through
\begin{equation}
\mathcal L=\mathcal L_{\mathrm q}
+\lambda_{\mathrm{rank}}\mathcal L_{\mathrm{rank}}
+\lambda_{\mathrm{safe}}\mathcal L_{\mathrm{safe}}
+\lambda_{\mathrm{coh}}\mathcal L_{\mathrm{coh}}.
\label{eq:loss}
\end{equation}
$\mathcal L_{\mathrm q}$ averages pinball loss over $D,R,A$, non-nominal candidates, and quantile levels, estimating the distributions needed for conservative choice. Pairwise logistic ranking on median $A$ emphasizes action ordering, and cross-entropy supervises adverse-event filtering. The coherence term $|\widehat Q_{A,k}^\tau-\widehat Q_{D,k}^\tau-\widehat Q_{R,k}^\tau|$ softly couples the heads during training. This is a training regularizer rather than an assumption of quantile additivity. Selection uses the separately predicted total quantiles, and Q4 evaluates the contribution of this coupling. Calibration follows training. Loss coefficients are specified with the experimental settings.

\input{experiments}

\section{Conclusion and Future Work}

We presented CARE, a framework that connects local predictive memory, explicit teammate-response supervision, and calibrated action selection. Matched reactive and replay continuations identify how teammate adaptation changes an action's value, and the analysis characterizes when that contribution changes the preferred decision. With representation, proposals, and data fixed, response supervision improves ranking and regret over learning the same information through branch utilities. Results on RoboFactory, MARS, and two physical tasks demonstrate effective decentralized manipulation through a capable nominal policy and selective coordination adjustments. Future work can adapt proposal distributions~\cite{long2026casf}, extend response learning to changing teammate policies, and combine selection with runtime monitoring~\cite{cheng2025dose3,zhou2026patch}.

\bibliographystyle{IEEEtran}
\bibliography{references}

\end{document}

%% file: authors.tex
\hypersetup{pdfauthor={Zihao Li, Yanan ZHOU, Ranpeng Qiu, Guoqiang Ren, Weiming Zhi}}
\author{%
    Zihao Li\textsuperscript{1,2},
    Yanan ZHOU\textsuperscript{1},
    Ranpeng Qiu\textsuperscript{2},
    Guoqiang Ren\textsuperscript{2},
    Weiming Zhi\textsuperscript{1,*}\\[0.4em]
    {\normalsize\textsuperscript{1}The University of Sydney
    \qquad\textsuperscript{2}Zeno AI}%
    \thanks{*Corresponding author: Weiming Zhi.\newline
    \href{mailto:weiming.zhi@sydney.edu.au}{\texttt{weiming.zhi@sydney.edu.au}}.}%
}

%% file: figure_care_labels.tex
\begingroup
\definecolor{careencoderfill}{RGB}{232,239,255}
\definecolor{carereadoutfill}{RGB}{234,243,240}
\begin{tikzpicture}
  \node[anchor=south west,inner sep=0] (careimage) at (0,0)
    {\includegraphics[width=0.9\linewidth]{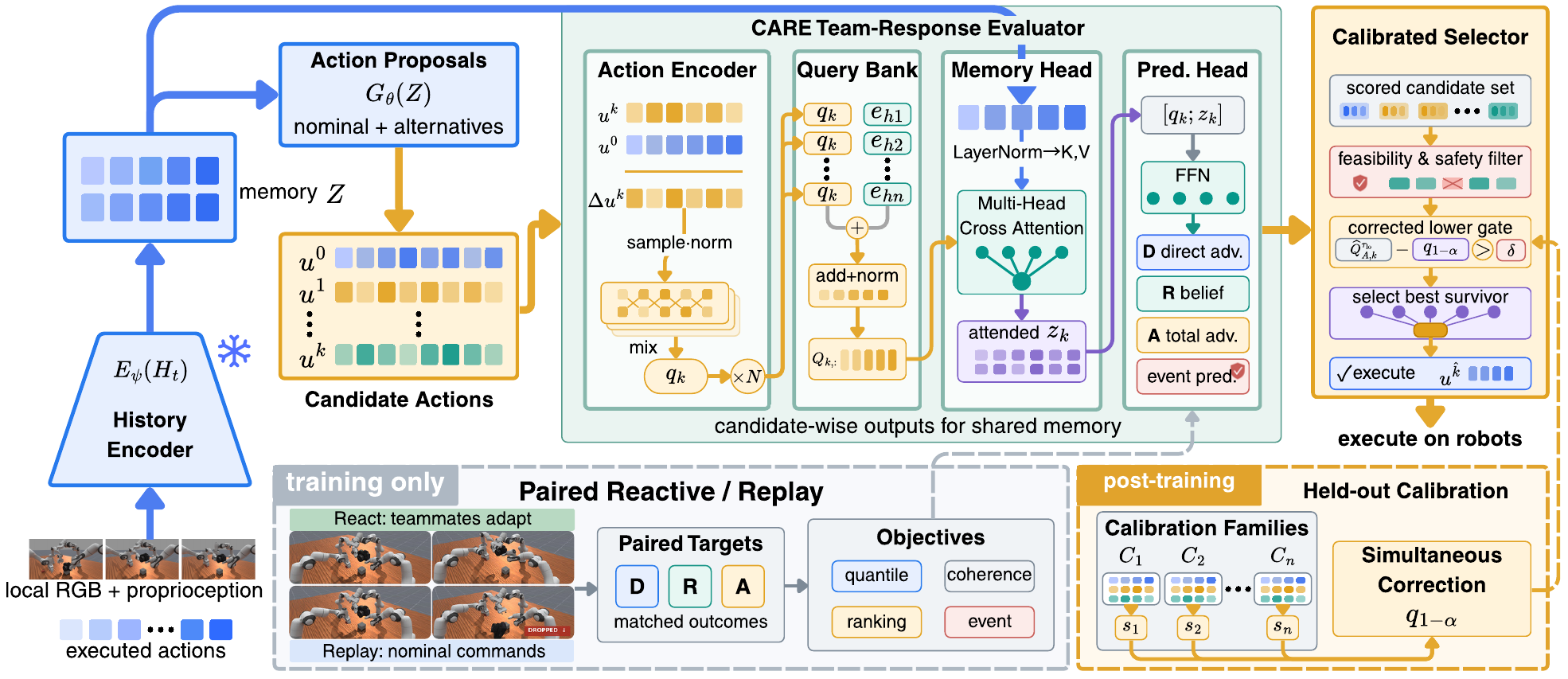}};
  \begin{scope}[x={(careimage.south east)},y={(careimage.north west)}]
    \fill[careencoderfill] (0.057,0.321) rectangle (0.132,0.407);
    \node[align=center,inner sep=0,font=\sffamily\bfseries\fontsize{6.4}{7.4}\selectfont]
      at (0.0945,0.365) {Predictive\\belief encoder};
    \fill[carereadoutfill] (0.468,0.942) rectangle (0.733,0.980);
    \node[inner sep=0,font=\sffamily\bfseries\fontsize{8}{9}\selectfont]
      at (0.605,0.963) {CARE response readout};
    \fill[careencoderfill] (0.219,0.841) rectangle (0.299,0.889);
    \node[inner sep=0,font=\fontsize{8}{9}\selectfont]
      at (0.258,0.865) {$u^0(H,Z_b)$};
    \fill[careencoderfill] (0.194,0.793) rectangle (0.317,0.829);
    \node[inner sep=0,font=\sffamily\fontsize{6.6}{7.2}\selectfont]
      at (0.255,0.811) {nominal + transforms};
  \end{scope}
\end{tikzpicture}
\endgroup

%% file: experiments.tex
\section{Experiments}
\label{sec:experiments}

\begin{figure*}[!t]
\centering
\includegraphics[width=\textwidth]{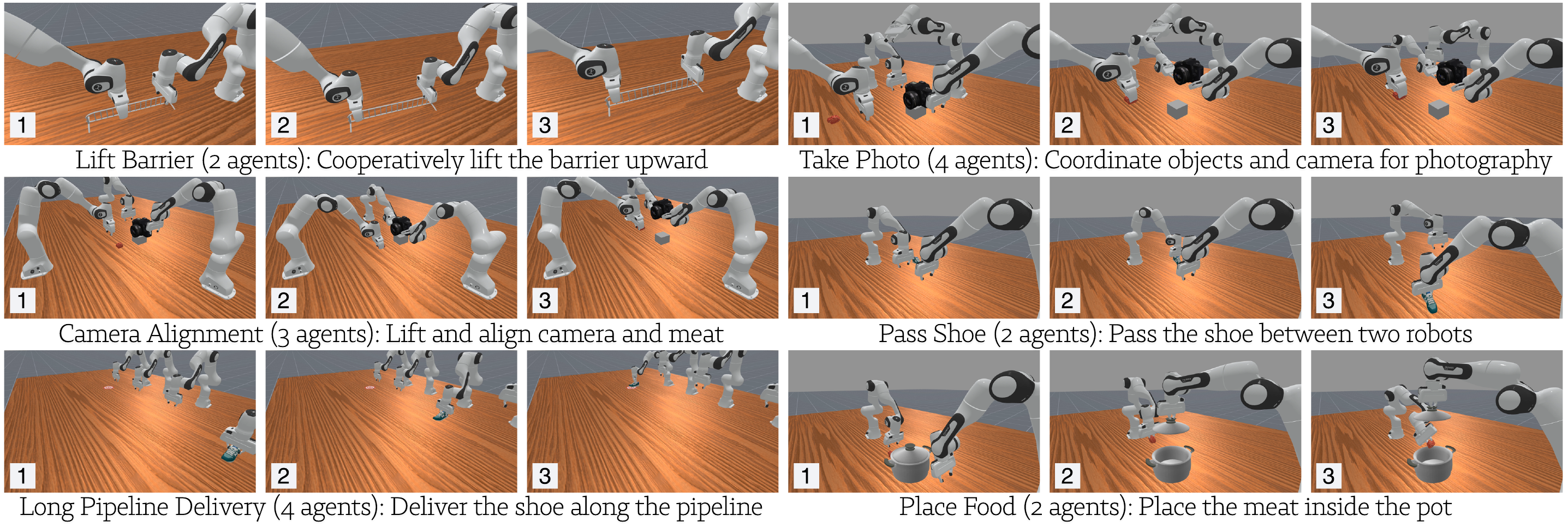}
\vspace{-1.5em}
\caption{\textbf{Collaborative tasks in RoboFactory.} Representative sequences of the six tasks with teams of two to four robots.}
\label{fig:sim-result}
\end{figure*}

\begin{table}[t]
\caption{Closed-loop SR (\%) on RoboFactory and the MARS Control Track.}
\label{tab:main-comparison}
\vspace{-0.5em}
\centering
\studytable
\scriptsize
\setlength{\tabcolsep}{2.2pt}

\begin{tabular*}{\columnwidth}{@{\extracolsep{\fill}}lrrrrrrr@{}}
\toprule
\multicolumn{8}{c}{\textbf{RoboFactory}} \\
\cmidrule(lr){1-8}
Policy & LB\nrob{2} & CA\nrob{3} & LP\nrob{4} & TP\nrob{4} &
PS\nrob{2} & PF\nrob{2} & Avg. \\
\midrule
\multicolumn{8}{l}{\textit{No external robot-action pretraining}} \\
ACT \cite{zhao2023act}
    & 90 & 50 & 80 & 0 & 5 & \textbf{100} & 54.17 \\
Diffusion Policy \cite{chi2023diffusion}
    & 65 & 20 & 0 & 5 & 0 & 35 & 20.83 \\
LatentToM$^{\dagger}$ \cite{he2025latenttom}
    & \textbf{100} & 35 & \textbf{95} & 35 & 95 & 75 & 72.50 \\
GauDP$^{\dagger}$ \cite{wang2025gaudp}
    & 70 & 10 & 0 & 0 & 0 & 70 & 25.00 \\
ManiFlow \cite{yan2025maniflow}
    & 75 & 50 & 0 & 40 & 20 & 70 & 42.50 \\
\midrule
\multicolumn{8}{l}{\textit{External robot-action pretraining}} \\
RDT-1B \cite{liu2024rdt}
    & 35 & 25 & 0 & 40 & 20 & 60 & 30.00 \\
OpenVLA-OFT \cite{kim2025oft}
    & 10 & 55 & 0 & 15 & 0 & 15 & 15.83 \\
$\pi_{0.5}$ \cite{black2025pi05}
    & 85 & \textbf{65} & 55 & 65 & 70 & 15 & 59.17 \\
\midrule
\multicolumn{8}{l}{\textit{CARE: visual pretraining only}} \\
\textbf{CARE (ours)}
    & \textbf{100} & 60 & 90 & \textbf{100}
    & \textbf{100} & 65 & \textbf{85.83} \\
\bottomrule
\end{tabular*}

\vspace{0.35em}

\begin{tabular*}{\columnwidth}{@{\extracolsep{\fill}}lrrrrrr@{}}
\toprule
\multicolumn{7}{c}{\textbf{MARS Control Track}} \\
\cmidrule(lr){1-7}
Policy & PC\nrob{2} & SC\nrob{2} & TS\nrob{3} & FS\nrob{4} &
Avg. & Mean \\
\midrule
\multicolumn{7}{l}{\textit{No external robot-action pretraining}} \\
ACT \cite{zhao2023act}
    & \textbf{30} & 45 & \textbf{15} & 0 & 22.50 & 38.33 \\
Diffusion Policy \cite{chi2023diffusion}
    & 0 & 0 & 0 & 0 & 0.00 & 10.42 \\
LatentToM$^{\dagger}$ \cite{he2025latenttom}
    & \textbf{30} & 35 & 0 & 0 & 16.25 & 44.38 \\
GauDP$^{\dagger}$ \cite{wang2025gaudp}
    & 0 & 0 & 0 & 0 & 0.00 & 12.50 \\
ManiFlow \cite{yan2025maniflow}
    & 25 & 35 & 0 & 0 & 15.00 & 28.75 \\
\midrule
\multicolumn{7}{l}{\textit{External robot-action pretraining}} \\
RDT-1B \cite{liu2024rdt}
    & 15 & 35 & 10 & 0 & 15.00 & 22.50 \\
OpenVLA-OFT \cite{kim2025oft}
    & 0 & 0 & 0 & 0 & 0.00 & 7.92 \\
$\pi_{0.5}$ \cite{black2025pi05}
    & 15 & 55 & \textbf{15} & 0 & 21.25 & 40.21 \\
\midrule
\multicolumn{7}{l}{\textit{CARE: visual pretraining only}} \\
\textbf{CARE (ours)}
    & 15 & \textbf{95} & 10 & \textbf{20}
    & \textbf{35.00} & \textbf{60.42} \\
\bottomrule
\end{tabular*}

\studynote{Subscripts denote robot counts.
$^{\dagger}$Run under the shared decentralized interface.
LB Lift Barrier, CA Camera Alignment, LP Long Pipeline Delivery,
TP Take Photo, PS Pass Shoe, PF Place Food,
PC Place Cube in Cup, SC Strike Cube Hard,
TS Three Robots Place Shoes, FS Four Robots Stack Cube.}
\vspace{-1em}
\end{table}
We evaluate how CARE combines predictive memory, response learning, and calibrated selection to coordinate decentralized manipulation. Q1 and Q6 assess the complete policy in simulation and on hardware. Q2 identifies the decisions that benefit from teammate response, Q3 and Q4 examine candidate selection and supervision, and Q5 studies when the policy should intervene. Together, these comparisons connect task performance to the decisions underlying it.

\vspace{-0.5em}

\subsection{Benchmarks and Protocol}


We evaluate six collaborative RoboFactory tasks~\cite{qin2025robofactory} and four hidden-seed MARS Control Track tasks~\cite{kang2026mars}, including teams of three and four robots (Fig.~\ref{fig:sim-result}). Within each benchmark, policies use the same demonstrations, local observation boundary, action interface, control rate, episode limits, and shared arm weights. Each CARE robot runs the same frozen $\pi_{\mathrm{CARE}}$ from its own local history, while label generation uses the shared reference policy $\pi_{\mathrm{ref}}$ defined below and baselines retain their own policies. Only the external-pretraining group may use robot-action data outside the benchmark. At execution, each arm observes only its RGB history, proprioception, executed-action history, and task instruction, never teammate observations or simulator state.

\textbf{Evaluation objective.} Team utility measures task progress and completion together with coordination costs under a fixed scalarization. Progress is clipped to $[-1,1]$, and completion indicates success within the horizon. Conflict, duplicate work, and deadlock are measured as fractions of control steps. Activity imbalance is the range of robot activity fractions, while delay is time to success divided by the horizon, or one if unsuccessful. Progress and completion contribute positively, and the remaining terms are costs. Collisions and drops supply a separate adverse-event label. All response variants share this objective and its scale.

\textbf{Nominal policy and predictive representation.} Frozen DINO features~\cite{caron2021dino}, proprioception, executed actions, and task instruction form a 16-step local history. A temporal Transformer and action-query decoder predict joint-target chunks~\cite{zhao2023act}. The predictive belief encoder uses causal attention and discrepancy-based event memory, with ego, teammate, and interaction slots. Its belief-conditioned residual follows \eqref{eq:belief-proposal}. Predictive training uses future visual features at 0.2, 0.4, 0.8, and 1.6 seconds, teammate state and action targets, and teacher reconstruction. Regularization includes exchange consistency, a feature-variance floor, and action ranking under matched versus shuffled belief representations. These demonstration-based objectives are distinct from supervision from reactive and replay branches. DINO supplies visual pretraining only.

\textbf{Training order and reference policy.} Demonstration training uses team and future information for supervision, while execution uses only local history. We freeze the nominal path as $\pi_{\mathrm{ref}}(H)=u^0(H)$, retaining its predictive belief encoder. Paired continuations match physical state, observation history, policy memory, pending commands, and random-number state. Reactive teammates recompute the nominal action, whereas replay uses recorded commands. All response variants share one frozen dataset and reference policy. Only the unmatched control changes pairing. Candidate differences and horizon embeddings query the frozen belief and event representation by cross-attention. We train the shared readout and calibrate selection, with no iterative relabeling. World-model augmentation supplies observations after altered actions, but response targets still require both matched branch outcomes. The complete $\pi_{\mathrm{CARE}}$ runs independently on every robot.

\textbf{Settings.} CARE scores $K=6$ candidates over a 16-step horizon. Besides $u^0$, five transforms are held over the 8-step commitment window, namely waiting for 4 steps, retreating by half the nominal displacement, advancing or delaying the gripper by 4 steps, and slowing motion to $0.6\times$. Quantile levels are $\mathcal T=\{.05,.25,.5,.75,.95\}$ with $\tau_{\mathrm{lo}}=.05$. Calibration uses $M_{\mathrm{cal}}=24$ families at $1-\alpha=.9$, with $\delta=0$, $\eta=.25$, and loss weights $(.20,.10,.10)$ in \eqref{eq:loss}. The seven utility components listed above use relative weights $(30,30,8,6,6,3,5)$, normalized by their sum. Each family contains two repeats of $K$ reactive and $K$ replay continuations of at most 64 steps, from which all horizons are read, giving at most 1,536 simulated control transitions per family and about 276,000 over 180 families. This budget covers simulator labeling rather than world-model augmentation, and deployment uses neither simulator nor world-model rollouts.

\textbf{Metrics.} SR is terminal success over 20 rollouts per task, equally weighted, and OR is the percentage of control steps replacing $u^0$. Offline, headroom is the reactive oracle's realized advantage over $u^0$, Recov.\ the fraction recovered, Regret the gap to the best candidate, and Rank the accuracy of candidate ordering against paired reactive outcomes. $R$ MAE is the mean absolute residual of a variant's response predictor on held-out paired labels (Q4). Headroom, Regret, $R$ MAE, and $\Gamma$ use the common utility scale defined above.

\textbf{Splits and statistics.} The snapshot evaluation uses 24 held-out test families. Table~\ref{tab:mechanism-ablation} reports the deployed seed. Figure~\ref{fig:evaluator-stats} instead summarizes three seeds paired by family and seed. We also use trajectory-grouped five-fold cross-validation over all 180 families, with calibration excluded from the test trajectory. Confidence intervals use 10,000 bootstrap resamples clustered by trajectory and family, keeping snapshots, candidates, repeats, and both branches from one trajectory together. Table~\ref{tab:response-strata} reports grouped-CV stratum averages. SR in the mechanism tables is a 120-rollout consistency check. Arrows mark the preferred direction and bold the best entry.

\vspace{-0.5em}

\begin{table*}[!t]
\caption{Response strata on the 180 held-out decisions under trajectory-grouped cross-validation.}
\vspace{-0.5em}
\label{tab:response-strata}
\centering
\studytable
\setlength{\tabcolsep}{3.5pt}
\begin{tabular*}{\textwidth}{@{\extracolsep{\fill}}lr rrr rrrr rrrr@{}}
\toprule
 & & \multicolumn{3}{c}{$\Gamma$ (utility units)} & \multicolumn{4}{c}{Recov. $\uparrow$} & \multicolumn{4}{c}{Regret $\downarrow$} \\
\cmidrule(lr){3-5}\cmidrule(lr){6-9}\cmidrule(lr){10-13}
Stratum & $n$ & Mean & Median & P90 & Direct & Total & Branches & CARE & Direct & Total & Branches & CARE \\
\midrule
Response-dependent & 108 & 0.0124 & 0.0108 & 0.0240 & 0.060 & 0.390 & 0.450 & 0.560 & 0.0076 & 0.0059 & 0.0054 & 0.0046 \\
\quad positive reversal & 68 & 0.0148 & 0.0131 & 0.0265 & 0.040 & 0.440 & 0.500 & 0.630 & 0.0084 & 0.0054 & 0.0049 & 0.0039 \\
\quad negative reversal & 40 & 0.0083 & 0.0074 & 0.0178 & 0.090 & 0.305 & 0.365 & 0.440 & 0.0062 & 0.0068 & 0.0063 & 0.0058 \\
Response-invariant & 72 & 0 & 0 & 0 & 0.000 & 0.000 & 0.000 & 0.000 & 0.0000 & 0.0012 & 0.0010 & 0.0009 \\
\midrule
All decisions & 180 & 0.0074 & 0.0065 & 0.0180 & 0.036 & 0.234 & 0.270 & 0.336 & 0.0046 & 0.0040 & 0.0036 & 0.0031 \\
\bottomrule
\end{tabular*}
\studynote{Direct, Total, and Branches denote the evaluators of Table~\ref{tab:mechanism-ablation} under CARE's proposals and selection rule.}
\vspace{-0.5em}
\end{table*}

\subsection{Q1. How Capable Is CARE?}
\label{sec:main-closed-loop}


Table~\ref{tab:main-comparison} compares CARE with action-only policies and the collaboration-specific LatentToM and GauDP. The pretraining partition separates external robot-action pretraining from visual-only pretraining. This comparison evaluates complete training recipes, including CARE's paired interaction supervision. Q4 then controls the available supervision to examine the contribution of its response targets.

On RoboFactory, CARE achieves the highest average success. Its strongest results occur on Take Photo and Pass Shoe, where holding or waiting for a partner is central to completion. CARE succeeds in every rollout on both tasks. LatentToM is the strongest baseline and also performs well on Pass Shoe, reinforcing the importance of anticipating partner behavior. Precision remains a complementary requirement, as other policies lead on Camera Alignment, Long Pipeline Delivery, and Place Food. Selection still depends on the execution accuracy of available actions. CARE's nominal path already achieves 80.8\% success, and response-based selection raises it to 85.8\% with sparse overrides (Table~\ref{tab:selector-ablation}). Predictive memory supports the nominal policy, while the response readout identifies when a local adjustment can improve the team's continuation.

\begin{figure}[!t]
\centering
\includegraphics[width=\linewidth]{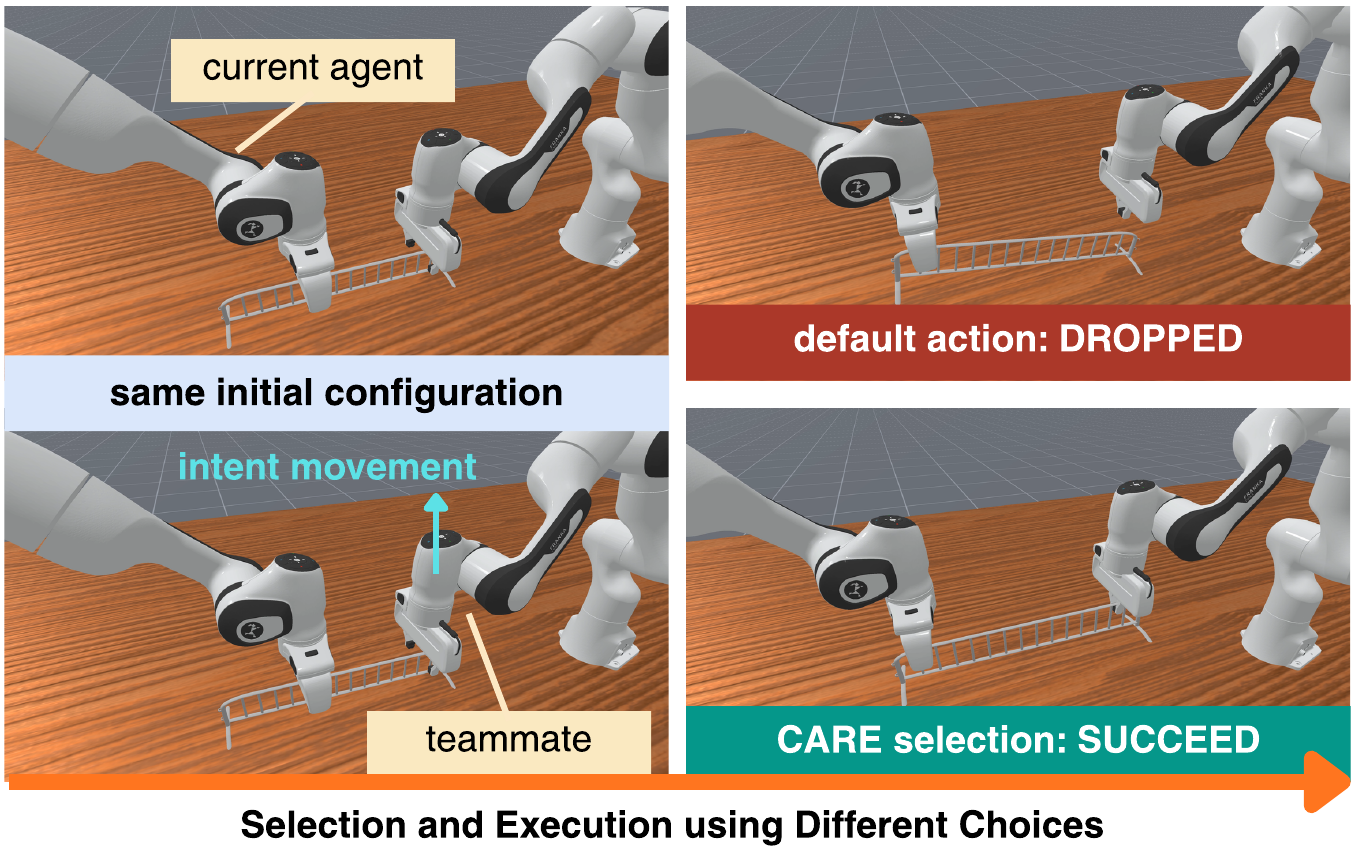}
\vspace{-1.5em}
\caption{\textbf{Teammate response matters for action selection.} From the same Lift Barrier configuration (left), the default action drops the barrier (top right), while CARE's response-aware selection follows the teammate's upward motion and completes the lift (bottom right).}
\label{fig:selection-result}
\vspace{-0.5em}
\end{figure}

On MARS, CARE also achieves the highest average success, led by Strike Cube Hard and the only successful trials on Four Robots Stack Cube. CARE trails the strongest baselines on the other two tasks, and the low success rates across methods show the difficulty of these settings. Together, the two benchmarks demonstrate that CARE's overall advantage extends across different collaborative tasks and team sizes.

\vspace{-0.5em}

\subsection{Q2. When Does Team Response Matter for Choice?}
\label{sec:response-criticality}


We characterize response relevance using $\Gamma(H)$ from \eqref{eq:response-threshold}. Candidate means are estimated by averaging two paired simulator repeats, and strata are frozen before selector comparison. \emph{Response-dependent} decisions have positive estimated $\Gamma$ beyond the evaluation tolerance. In \emph{response-invariant} decisions, a direct-only choice remains optimal within that tolerance. Positive reversals involve a candidate with $d_k\leq0<a_k$, while negative reversals involve one with $d_k>0\geq a_k$. The reversals distinguish helpful adaptation from disruptive responses.

Table~\ref{tab:response-strata} shows that teammate response changes the preferred action in 108 of the 180 held-out decisions. CARE has the highest recovery and lowest regret in this stratum, with its clearest advantage on positive reversals. Waiting or maintaining support can look unproductive locally while enabling a partner to finish its motion. Negative reversals test the complementary ability to avoid seemingly useful actions that disrupt coordination. Total-only prediction is less reliable here, whereas CARE again achieves the lowest regret. In response-invariant decisions, Direct is already optimal and CARE incurs a small residual regret. Zero recovery in this stratum reflects the absence of positive reactive headroom in these samples. On Lift Barrier (Fig.~\ref{fig:selection-result}), moving before the partner lifts drops the barrier, while CARE follows its upward motion and preserves support.

\begin{table}[!t]
\caption{Mechanism studies with fixed proposals.}
\vspace{-0.5em}
\label{tab:headroom-selection}
\label{tab:mechanism-ablation}
\centering
\studytable
\begin{tabularx}{\linewidth}{@{}Xrrrr@{}}
\toprule
Proposals / selector & SR $\uparrow$ & Recov. $\uparrow$ & Regret $\downarrow$ & OR \\
\midrule
Perturbed / CARE & 84.2 & 0.450 & 0.0055 & 6.5 \\
CARE / Random & 63.3 & 0.140 & 0.0086 & 96.7 \\
CARE / Direct-only & 80.0 & 0.040 & 0.0096 & 42.0 \\
CARE / Total-only & 81.7 & 0.280 & 0.0072 & 28.0 \\
\textbf{CARE / CARE} & \textbf{85.8} & \textbf{0.520} & \textbf{0.0048} & 6.0 \\
\midrule
Supervision variant & SR $\uparrow$ & $R$ MAE $\downarrow$ & Rank $\uparrow$ & Regret $\downarrow$ \\
\midrule
Total only ($A$) & 81.7 & \nr$^{\ast}$ & 0.548 & 0.0072 \\
Total + direct ($A+D$) & 83.3 & 0.0310$^{\dagger}$ & 0.571 & 0.0059 \\
Total + branch utilities & 85.0 & 0.0610$^{\dagger}$ & 0.603 & 0.0061 \\
Direct only ($D$) & 80.8 & \nr$^{\ast}$ & 0.452 & 0.0083 \\
Unmatched branches & 78.3 & 0.1290 & 0.438 & 0.0096 \\
CARE w/o coherence & 84.2 & 0.0520 & 0.587 & 0.0057 \\
\textbf{CARE (full)} & \textbf{85.8} & \textbf{0.0140} & \textbf{0.671} & \textbf{0.0048} \\
\bottomrule
\end{tabularx}
\studynote{Snapshot split, deployed seed. $^{\dagger}$Contrast of marginal medians. $^{\ast}$No response head (Q4).}
\vspace{-0.5em}
\end{table}

\begin{figure*}[!t]
\centering
\includegraphics[width=\linewidth]{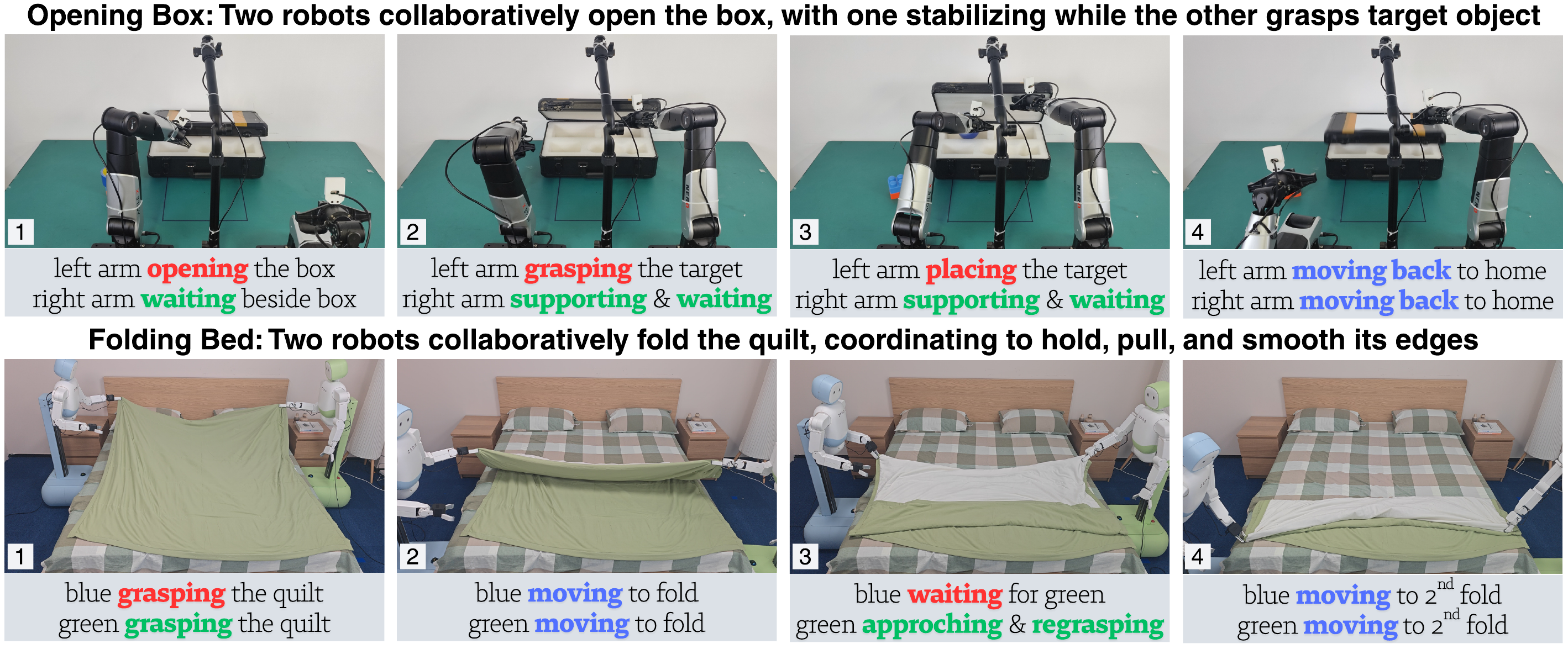}
\vspace{-1.5em}
\caption{\textbf{CARE on two physical collaboration tasks.} In \emph{Opening Box} (top), one arm supports the lid and waits while its partner grasps and places the object. In \emph{Folding Bed} (bottom), one arm holds the quilt edge and waits for the partner to regrasp. Keyframes highlight sustained support and waiting.}
\label{fig:real-robot-screenshot}
\vspace{-0.5em}
\end{figure*}

\begin{table}[!t]
\caption{Selection rules with the same trained response readout.}
\vspace{-0.5em}
\label{tab:selector-ablation}
\centering
\studytable
\begin{tabularx}{\linewidth}{@{}Xrr@{}}
\toprule
Selector & OR (\%) & SR (\%) $\uparrow$ \\
\midrule
Nominal path (selector off) & 0.0 & 80.8 \\
Posterior mean & 98.0 & 42.5 \\
Uncalibrated lower quantile & 38.0 & 78.3 \\
Candidate-wise calibration & 18.0 & \textbf{89.2} \\
Simultaneous calibration & 6.0 & 85.8 \\
\midrule
\textbf{CARE (+ safety filter)} & 6.0 & 85.8 \\
\bottomrule
\end{tabularx}
\vspace{-0.15\baselineskip}
\vspace{-0.5em}
\end{table}

\begin{figure}[t]
\centering
\includegraphics[width=\linewidth]{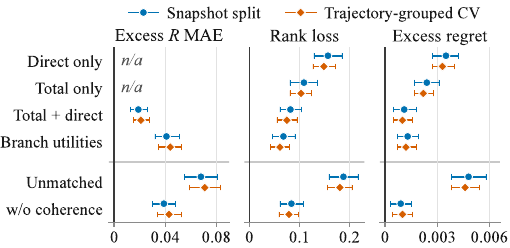}
\vspace{-1em}
\caption{\textbf{Response-readout ablations across three seeds.} Mean paired differences from CARE with 95\% bootstrap intervals on the snapshot split (circles) and trajectory-grouped CV (diamonds). Table~\ref{tab:mechanism-ablation} reports one deployed seed, not these three-seed means. Upper rows vary the supervision targets and lower rows the coupling.}
\label{fig:evaluator-stats}
\end{figure}

\subsection{Q3. Does the Response Readout Improve Selection?}

The upper block of Table~\ref{tab:headroom-selection} compares evaluators over shared candidates with the proposal model fixed. A separate proposal control uses first-action perturbations of $u^0$, including a one-step hold, time warps, and a frozen gripper. These brief changes provide less opportunity for adaptation than CARE's sustained behavioral alternatives.

Random selection wastes the candidates' concentrated headroom by often replacing a capable nominal action with an inferior alternative. Direct-only selection also intervenes often but recovers little gain, showing why immediate effects are insufficient for coordination. Total-only prediction improves recovery, yet CARE recovers roughly twice as much with far fewer overrides. CARE intervenes at about 6\% of steps while recovering half the available headroom. The weaker first-action perturbations support pairing sustained behavioral alternatives with a readout that evaluates their team benefit.

\begin{table}[t]
\vspace{-0.1\baselineskip}
\caption{Real-world dual-robot evaluation.}
\vspace{-0.5em}
\label{tab:real-robot}
\centering
\studytable
\setlength{\tabcolsep}{2.5pt}
\begin{tabularx}{\linewidth}{@{}Xrrrr@{}}
\toprule
Policy & FB & OB & SR $\uparrow$ & OR \\
\midrule
ACT~\cite{zhao2023act} & $12/20$ & $10/20$ & 55.0 & n/a \\
$\pi_{0.5}$~\cite{black2025pi05} & $9/20$ & $14/20$ & 57.5 & n/a \\
CARE, selector off & $18/20$ & $15/20$ & 82.5 & 0.0 \\
\textbf{CARE} & $20/20$ & $17/20$ & \textbf{92.5} & 5.8 \\
\bottomrule
\end{tabularx}
\studynote{FB Folding Bed, OB Opening Box. Successes out of 20 trials.}
\vspace{-0.2\baselineskip}

\end{table}

\subsection{Q4. Does Explicit Response Supervision Help?}

The lower block of Table~\ref{tab:mechanism-ablation} retrains the readout with proposals, representation, data, budget, and selection rule fixed. Total-only and $A+D$ use fewer targets from the same paired outcomes. The stronger branch-utility control predicts all four utilities and total advantage, providing every outcome needed to construct $R$. Unmatched branches remove seed coupling, while the coherence ablation removes the penalty linking the heads. The table reports the deployed seed, and Fig.~\ref{fig:evaluator-stats} averages paired differences across three seeds.

\textbf{Response targets improve action evaluation.} CARE improves ranking accuracy by $0.068$ and reduces regret by $0.0013$ relative to branch-utility supervision. Both gains persist in the three-seed snapshot comparison, whose paired confidence intervals exclude zero (Fig.~\ref{fig:evaluator-stats}). Both methods learn a total-advantage head, so response targets improve the selection score. Grouped cross-validation also favors CARE in ranking, and Table~\ref{tab:response-strata} locates its regret advantage in response-dependent decisions. Although branch utilities contain the same information, explicitly supervising adaptation helps the readout distinguish team value from direct effects. These controlled decision gains complement the closed-loop results.

\textbf{Pairing and coupling support response learning.} Removing seed coupling gives the weakest ranking and largest regret in the supervision study. Removing coherence also increases response error and worsens selection. These results support matching branch conditions and jointly learning the related targets. Here $R$ MAE measures $\E|g(H)-R_k|$, minimized by the conditional median. CARE and its unmatched and coherence variants use $\widehat Q^{.5}_{R,k}$, while $A+D$ and branch supervision use contrasts of predicted marginal medians. Medians do not commute with subtraction, so this diagnostic compares the resulting response estimators. Rank and Regret directly assess their decisions. Total-only and Direct-only have no response predictor.

\subsection{Q5. Can Selection Balance Intervention and Abstention?}

Table~\ref{tab:selector-ablation} holds CARE's readout fixed and varies the decision rule. Posterior-mean selection overrides almost continuously and sharply reduces success. An uncalibrated lower quantile still falls below nominal, motivating a calibrated threshold for intervention.

Calibration supplies this threshold. Candidate-wise calibration achieves the highest success, while simultaneous calibration retains a gain over nominal with one third as many overrides. Since the common shift preserves candidate order, calibration governs abstention and complements the ranking improvements in Q4. The safety filter leaves aggregate rates unchanged. All robots run the tested selector, so these gains include the effects of teammates departing from $\pi_{\mathrm{ref}}$. Sparse intervention retains the shared nominal policy at most control steps, providing a practical basis for reference-based evaluation. However, aggregate OR does not measure concurrent overrides or their downstream effects. The results establish effective collective execution, while isolating the policy mismatch requires varying which robots intervene and how strongly.

\subsection{Q6. How Does CARE Perform on Physical Tasks?}

We compare CARE with ACT and $\pi_{0.5}$ on \emph{Folding Bed} and \emph{Opening Box} (Fig.~\ref{fig:real-robot-screenshot}), with 20 trials per task and policy. Policies share local observations, an 8-D joint-target action space, control frequency, and task budget. Policies are trained on task demonstrations and frozen. CARE arms share a checkpoint but use their own histories. The selector-off control retains the encoder and nominal path. Response supervision uses simulated-twin continuations, with a world model generating observations after altered actions. Calibration uses held-out simulator families.

Table~\ref{tab:real-robot} shows that CARE completes 37 of 40 trials, achieving the highest success across the two physical tasks. The nominal path already outperforms ACT and $\pi_{0.5}$, and selection adds four successful trials. Overrides are sparse and occur during holding and waiting phases. Supporting the quilt edge lets the partner regrasp, while holding the box lid preserves access to the object. Maintaining support thus enables the partner's progress.

The physical results complement Q3--Q5 by demonstrating coordinated execution with all selectors active. Execution requires one additional readout forward pass per control query and no simulator or world-model rollouts.